\documentclass[sigconf]{acmart}
\AtBeginDocument{%
  }

\usepackage{booktabs}
\usepackage{array}
\usepackage{amsmath}
\usepackage{mathtools}
\usepackage{multirow}
\usepackage{enumitem}
\usepackage{listings}
\graphicspath{{figures/}}

\copyrightyear{2026}
\acmYear{2026}
\setcopyright{cc}
\setcctype{by}
\acmConference[CAIS '26]{ACM Conference on AI and Agentic Systems}{May 26--29, 2026}{San Jose, CA, USA}
\acmBooktitle{ACM Conference on AI and Agentic Systems (CAIS '26), May 26--29, 2026, San Jose, CA, USA}
\acmDOI{10.1145/3786335.3813159}
\acmISBN{979-8-4007-2415-2/2026/05}

\begin{document}

\title{TraceFix: Repairing Agent Coordination Protocols with TLA+ Counterexamples}

\author{Shuren Xia}
\orcid{0009-0007-7459-7080}
\email{shuren.xia@rutgers.edu}
\affiliation{%
  \institution{Rutgers University}
  \city{Piscataway}
  \state{NJ}
  \country{USA}
}

\author{Qiwei Li}
\orcid{0009-0003-4013-4097}
\email{qiwei.li@rutgers.edu}
\affiliation{%
  \institution{Rutgers University}
  \city{Piscataway}
  \state{NJ}
  \country{USA}
}

\author{Taqiya Ehsan}
\orcid{0009-0007-1300-5379}
\email{taqiya.ehsan@rutgers.edu}
\affiliation{%
  \institution{Rutgers University}
  \city{Piscataway}
  \state{NJ}
  \country{USA}
}

\author{Jorge Ortiz}
\orcid{0000-0003-3325-1298}
\email{jorge.ortiz@rutgers.edu}
\affiliation{%
  \institution{Rutgers University}
  \city{Piscataway}
  \state{NJ}
  \country{USA}
}

\renewcommand{\shortauthors}{Xia et al.}

\begin{abstract}
We present TraceFix, a verification-first pipeline for Large Language Model (LLM) multi-agent coordination.
An agent synthesizes a protocol topology as a structured intermediate representation (IR) from a task description, generates PlusCal coordination logic, and iteratively repairs the protocol using counterexamples from the TLA+ model checker (TLC) until verification succeeds.
Verified process bodies are compiled into per-agent system prompts and executed under a runtime monitor that rejects out-of-topology coordination operations.
On 48 tasks spanning 16 scenario families, all tasks reach full TLC verification; 62.5\% pass on the first attempt and none requires more than four repair iterations.
State spaces span six orders of magnitude yet verification completes in under 60\,s for every task.
A 3{,}456-run runtime comparison shows that topology-monitored execution achieves the highest task completion (89.4\% average, 81.5\% full) and that runtimes using the verified protocol degrade at roughly half the rate of prompt-only and chat-only baselines when model capability is reduced.
A paired ablation under a fixed runtime shows that TLC-verified protocols cut deadlock/livelock (DL/LL) from 31.1\% to 14.1\%, with the largest separation under fault injection.
\end{abstract}



\maketitle

\section{Introduction}
Independent Large Language Model (LLM) agents increasingly run concurrently, communicate asynchronously, and mutate shared external state.
In this setting, coordination, instead of individual agent capability, is often the dominant source of system failures~\cite{cemri2025why}; races, deadlocks, missed handshakes, and premature termination emerge from interleavings and agent choices that are rarely exercised during design.
\textbf{TraceFix} addresses this problem with a verification-first protocol loop that turns model-checking counterexamples into actionable debugging signal for agent coordination protocols.

A natural response to coordination in concurrent multi agent systems is to ``add locks'' around shared resources.
In practice this reshapes failures rather than removing them.
Underspecified lock discipline trades races for deadlocks, and message passing introduces symmetric hazards when agents wait on messages that will never be sent.
These bugs are latent because they depend on rare schedules; a protocol can succeed in every observed run yet stall under an untested interleaving.
Centralized workflow engines avoid many such hazards by construction, but independent-agent multi-agent systems (MAS) inherit the concurrency hazards of distributed systems unless the coordination protocol itself is designed and verified to be safe~\cite{lamport1982byzantine,raju2019online,gonczarowski2012timely}.

Prior systems address LLM coordination through orchestration frameworks (explicit graphs, centralized scheduling)~\cite{chase2022langchain,li2023taskmatrix}, solver-in-the-loop generation (satisfiability modulo theories (SMT) feedback for single-artifact correctness)~\cite{ni2023lever,li2022alphacode}, or runtime guards (policy enforcement over action trajectories)~\cite{bai2022constitutional,rebedea2023nemo}.
TraceFix differs by targeting \emph{concurrent protocol correctness} rather than single-agent plans or workflow DAGs.
We introduce an explicit protocol topology intermediate representation (IR) plus a PlusCal~\cite{lamport2009pluscal} behavioral layer, a counterexample-driven repair loop that attacks the protocol until the TLA+ model checker (TLC)~\cite{lamport-tla} finds no violation, compilation of verified process bodies into prompts, and topology-aware runtime monitoring that rejects out-of-topology coordination operations.
To the best of our knowledge, this combination is new; no prior work threads verification through protocol design, repair, and runtime enforcement for LLM multi-agent coordination.

Writing a correct concurrent protocol is difficult for the same reason writing correct concurrent code is difficult: the hard cases live in interleavings that designers do not enumerate.
This limitation applies equally to LLM-generated protocols.
A one-shot specification, whether written by a human or an LLM, tends to be plausible but is underspecified because it lacks a mechanism to systematically discover edge cases.
TraceFix, therefore, treats the initial protocol as a hypothesis to be attacked.
It generates a model-checkable protocol candidate from a task description, uses exhaustive model checking under bounded assumptions to surface minimal counterexample traces, and repairs the protocol until the checker can no longer find violations.
After several iterations, the specification is no longer an approximation but the minimal protocol that survives an exhaustive search over all interleavings and agent choices \emph{under the chosen bounds}.

TraceFix represents protocols in two layers to make this loop reliable.
It first synthesizes a \textbf{protocol topology} as a structured IR that declaratively defines agents, shared resources, and directed labeled channels, and validates it before compilation.
It then generates a \textbf{PlusCal} scaffold from the topology and has the agent fill in per-agent process bodies (step-by-step PlusCal code that defines each agent's coordination protocol) as explicit coordination code (send, receive, acquire, release, and nondeterministic \texttt{either/or} branches).
PlusCal makes decision points explicit, allowing TLC to explore both scheduling interleavings and protocol-level choices.
After verification, TraceFix produces a \emph{candidate protocol} and a \emph{structured coordination interface} (locks, channels, labels); process bodies are compiled into prompts, and a runtime monitor enforces topology conformance by rejecting out-of-topology operations.
Step-order discipline is primarily carried out by prompts today~\cite{wei2022chainofthought,yao2022react}; the monitor enforces the interface, not the verified state machine (Section~\ref{sec:limitations}).

\paragraph{Scope.}
TraceFix targets coordination correctness, not semantic task correctness.
The pipeline checks safety-centric properties such as deadlock freedom (via TLC's \texttt{NoDeadlock} check), mutual exclusion over locks, and coordination soundness at termination (e.g. no orphan locks and drained channels), within a bounded model.
Task semantics and output quality are not verified, and liveness properties such as guaranteed termination or starvation freedom, are not checked.

\paragraph{Contributions.}
(1)~TraceFix, a protocol-design and verification agent that synthesizes a topology IR, generates a PlusCal coordination program, and uses TLC counterexamples for iterative repair until verification succeeds.
(2)~A coordination benchmark of 48 tasks spanning 16 scenarios at three difficulty tiers, designed so that ground truth is primarily coordination correctness rather than domain semantics.
(3)~A runtime topology enforcement layer that preserves agent autonomy while rejecting out-of-topology coordination operations.
(4)~A paired ablation on verified and unverified protocol variants that isolates the contribution of counterexample-driven repair under identical runtime conditions.

\paragraph{Empirical results.}
On a benchmark of 48 tasks spanning 16 scenario families, the pipeline achieves 100\% final TLC verification; 62.5\% of tasks pass on the first attempt, and no task requires more than four repair iterations.
Bounded model checking remains tractable (median under 1s, all tasks under 60s) even for protocols whose state space reaches millions of distinct states.
A controlled comparison of four runtime architectures under two model tiers (3{,}456 runs) shows that topology-monitored execution achieves the highest task completion (89.4\% average, 81.5\% full) and that runtimes using the verified protocol degrade at roughly half the rate of prompt-only or chat-only baselines when model capability is reduced.
A paired ablation under the Topology-monitored runtime shows that verified protocols reduce deadlock/livelock (DL/LL) from 31.1\% to 14.1\%, with the largest gap under injected faults.

\begin{figure*}[t]
    \centering
    \includegraphics[width=0.9\textwidth]{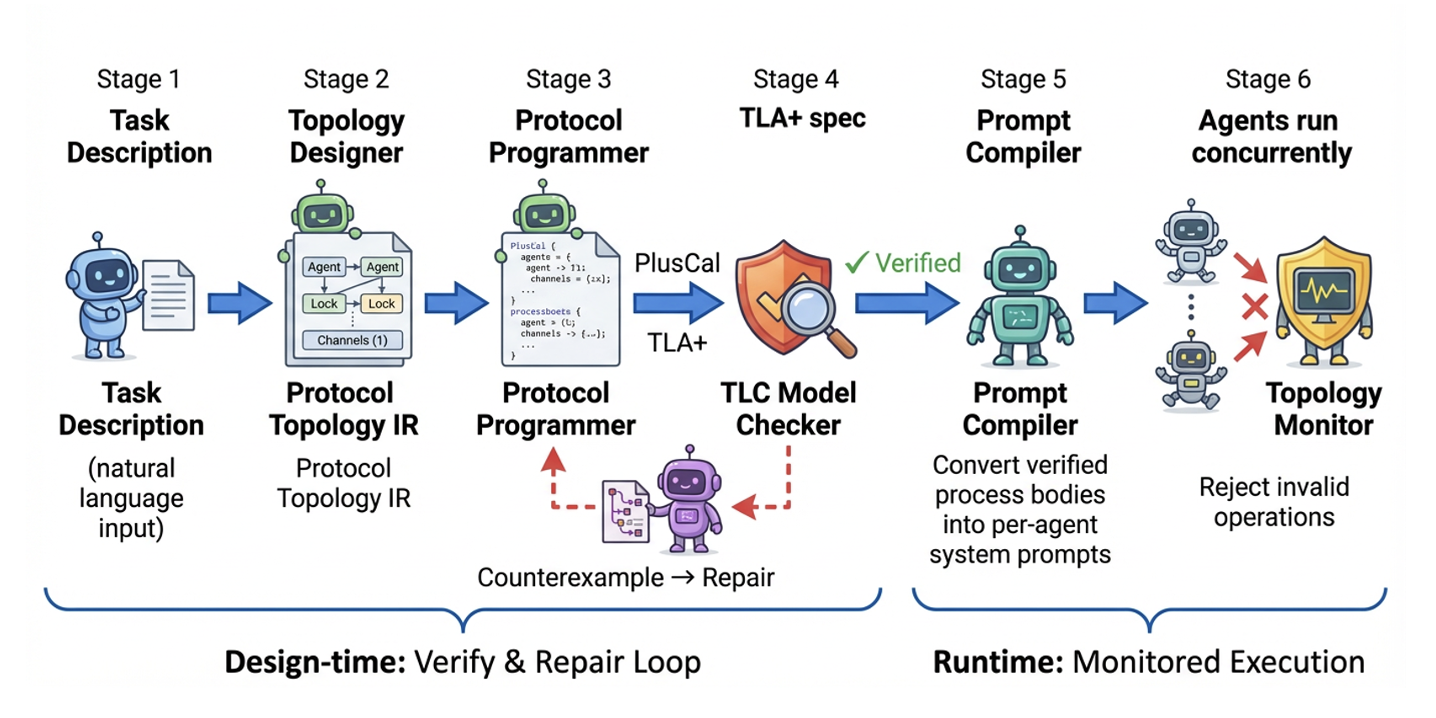}
    \caption{TraceFix pipeline overview. At design time (Stages 1--4), an orchestration agent synthesizes a protocol topology IR, generates PlusCal coordination logic, and iteratively repairs the protocol using TLC counterexamples until verification succeeds. At runtime (Stages 5--6), verified process bodies are compiled into per-agent prompts and executed under a topology monitor that rejects out-of-protocol coordination operations.}
    \label{fig:pipeline}
  \end{figure*}
\section{Method}

\subsection{A verification-first loop}
Concurrent protocols are difficult to get right because correctness depends on interleavings designers rarely enumerate; edge cases emerge only under untested schedules.
TraceFix treats protocol design as an iterative process with a tight feedback signal.

The pipeline has four stages.
First, an orchestration agent synthesizes a coordination protocol from a natural-language task description.
Second, the protocol is compiled into a model-checkable specification (PlusCal, then TLA+).
Third, TLC exhaustively searches for counterexamples under bounded assumptions, exploring all interleavings and nondeterministic choices within the chosen bounds.
Fourth, when TLC finds a violation, the repair agent revises the protocol based on the counterexample trace.
The loop terminates when TLC can no longer find a violation under the chosen bounds.

This design exploits a fundamental asymmetry; generating a correct concurrent protocol is hard, but checking one is mechanical.
The counterexample provides a concrete, minimal witness that pinpoints the exact interleaving responsible for the failure, so repairs are informed by evidence rather than guesswork.
The same loop has been used for protocol completion and synthesis~\cite{alur2015completion,egolf2024scythe}; TraceFix adapts it for LLM-synthesized agent coordination and connects it to runtime enforcement.

\subsection{Protocol design agent}
The orchestration agent plays three roles; \emph{topology designer}, \emph{protocol programmer}, and \emph{repairer}.
Given a task description and a structured guidance prompt, it first produces a \emph{protocol topology}, a declarative specification of who participates and how they may coordinate.
The topology fixes the set of agents, the shared resources (locks, optional counters), and the directed message channels with an explicit label vocabulary.
It then writes the coordination logic itself as PlusCal processes, one process per agent, with explicit control flow over sends, receives, and resource acquisition.

The separation of topology from behavior is deliberate.
The topology is declarative and structurally validatable; it captures the coordination envelope without requiring reasoning about interleavings.
Behavior is expressed in PlusCal with explicit control flow, making every decision point visible to the model checker.
When TLC finds a counterexample, repairs are applied to the PlusCal source, not the topology, because the topology defines the coordination interface while the behavior is where interleaving-sensitive hazards reside.

\subsection{Protocol topology; agents, resources, and channels}
The protocol topology is a compact IR that captures coordination structure before any behavioral logic is written.
It defines three kinds of elements.
\textbf{Agents} are the participants in the protocol.
\textbf{Resources} are shared coordination primitives; typically exclusive locks, with optional counters for pooled capacity (e.g., available slots).
\textbf{Channels} are directed first-in, first-out (FIFO) message queues with an explicit label vocabulary, defining which messages may be sent from which agent to which.
The topology induces a contract; it specifies which agent may send or receive on which channel, which resources exist, and how they may be acquired.

Before translation or model checking, the topology is validated structurally using a JavaScript Object Notation (JSON) schema and semantically; all referenced agents, resources, and channels must exist, channel directions must be respected, and label sets must be explicit.
This structural gate filters out malformed inputs at negligible cost, eliminating an entire class of downstream failures before the checker is ever invoked.

\subsection{PlusCal coordination logic}
Each agent is implemented as a PlusCal \texttt{fair process} whose body is a sequence of labeled steps.
Steps call coordination macros such as \texttt{send}, \texttt{receive}, \texttt{acquire\_lock}, and \texttt{release\_lock}, plus optional counter operations.
The \texttt{fair process} declaration embeds weak fairness in the generated specification; because only safety properties are checked (Section~\ref{sec:limitations}), the fairness conjunct does not affect verification results and is retained by PlusCal convention.

Control flow uses standard PlusCal constructs (\texttt{if}, \texttt{while}) plus \texttt{either/or} for nondeterministic choice.
The \texttt{either/or} construct is central to making LLM decision points verifiable; TLC explores all branches as part of the state space.
A protocol that models ``the agent may approve or request revisions'' with \texttt{either/or} will have both outcomes verified against all possible interleavings of other agents.
Even if a particular LLM would choose one branch almost always, the checker ensures that both branches are safe under every schedule.

\subsection{Translation and TLC checking}
TLA+ is a formal specification language for concurrent and distributed systems; it models behavior as transitions between states.
In TraceFix, protocols are written in PlusCal (an algorithm-like notation) and translated via \texttt{pcal.trans} into TLA+, yielding a transition-system specification suitable for model checking.
TLC explores reachable states under finite bounds, checks safety properties, and returns a counterexample trace when a property fails.
We check coordination-centric safety properties---mutual exclusion for locks, absence of orphan locks on termination, channel drainage when all agents complete, basic type invariants, and deadlock freedom.
These properties are listed in Table~\ref{tab:properties} along with what the checker does \emph{not} guarantee at runtime.

Because the full state space can grow very large, per-channel queue-depth bounds ($B_c$) keep exploration tractable.
Within those bounds, TLC explores all interleavings and all nondeterministic branches; nothing is sampled or approximated.
The bounds are enforced as invariants, so any execution that would exceed a queue bound is reported as a violation rather than silently truncated.

\paragraph{Property coverage rationale.} 
The MAST (multi-agent systems taxonomy)~\cite{cemri2025why} identifies specification and orchestration gaps as the dominant multi-agent failure mode.
Our four properties directly target these categories: deadlock freedom catches missed handshakes and premature termination; mutual exclusion prevents resource races; orphan-lock and channel-drainage checks catch incomplete cleanup.

\subsection{Formalization}
We model a coordination protocol as a concurrent transition system induced by a topology and a PlusCal program.
The topology $\mathcal{T} \triangleq (\mathcal{A}, \mathcal{R}, \mathcal{C})$ defines a finite set of agents $\mathcal{A}$, shared resources $\mathcal{R}$ (e.g., locks), and directed labeled FIFO channels $\mathcal{C}$.
Each channel $c \in \mathcal{C}$ has endpoints $(\mathsf{src}(c), \mathsf{dst}(c)) \in \mathcal{A}\times\mathcal{A}$ and label vocabulary $\Sigma_c$.
The translated state vector includes program counters $\mathsf{pc}$, channel queues $Q_c$, and resource ownership $H$ (e.g., $H(r) \in \mathcal{A} \cup \{\bot\}$ for a lock $r$).

TLC checks the properties in Table~\ref{tab:properties}.
Per-channel bounds $B_c$ are enforced as invariants ($\square\, \forall c.\; |Q_c| \le B_c$) to keep exploration finite.

\begin{table}[t]
\centering
\caption{Properties checked by TLC.}
\label{tab:properties}
\footnotesize
\setlength{\tabcolsep}{3pt}
\begin{tabular}{@{}lp{2.2cm}p{2.1cm}@{}}
\toprule
Property & Encoding & Limitations \\
\midrule
Mutual exclusion & $H$ single-valued & Step order \\
Deadlock freedom & \texttt{NoDeadlock} & Livelock, starvation \\
Orphan locks & $\mathsf{AllDone} \Rightarrow H{=}\bot$ & None \\
Channel drainage & $\mathsf{AllDone} \Rightarrow Q_c{=}\langle\rangle$ & None \\
\bottomrule
\end{tabular}
\end{table}

\subsection{Counterexample-driven repair}
When TLC produces a counterexample, it provides a trace; a concrete interleaving that demonstrates the hazard.
We treat this trace as a minimal bug report and feed it to the repair agent alongside the current topology and PlusCal source.
The repair agent edits PlusCal processes to eliminate the failure; typical repairs include enforcing a lock order, adding a missing handshake, or restructuring a loop to eliminate cyclic waits.

Over iterations, the protocol becomes more precise because it is forced to account for each edge case the checker discovers.
The key property is that each repair is informed by a specific failure witness, not by heuristic guessing; this is what distinguishes the loop from trial-and-error regeneration.
Section~\ref{sec:example} illustrates the repair process on a concrete task.

\subsection{End-to-end example}
\label{sec:example}
Appendix~\ref{app:example-3h} walks through a complete pipeline iteration on Task~3H (Large Survey Paper Production, Hard; 7 agents, 6 locks, 17 channels).
TLC discovers a deadlock whose counterexample reveals a hub agent that terminates before a satellite agent completes its revision loop, cascading through three downstream agents.
After two repair iterations, re-checking explores 7.7\,M states with zero violations.

\subsection{From verified protocols to runtime}
After verification, each agent's PlusCal process body is embedded verbatim into a system prompt that lists the agent's allowed channels and locks from the topology.
The runtime enforces two kinds of constraints derived from the topology.

First, a \emph{protocol monitor} validates that each coordination operation (send, receive, acquire, release) targets a declared channel or resource with a valid label; operations not admitted by the topology are rejected immediately.
Second, shared resources are implemented as blocking locks, so mutually exclusive access is enforced structurally.
The monitor does not enforce step order or fairness; it only ensures that coordination operations stay within the verified topology.
The remaining deadlocks in the Topology-monitored condition (8.8\% of runs, Section~4.3) arose from step-order deviations that topology checks cannot catch; a lightweight per-agent program-counter extension could close this gap (Section~\ref{sec:limitations}).

Formally, the monitor defines a set of allowed coordination events $\mathcal{E}(\mathcal{T})$ from the topology (well-formed channel endpoints, declared resources, valid labels).
At runtime, the system emits an event trace $\tau$; the monitor enforces $\tau \in \mathcal{E}(\mathcal{T})^*$ online by rejecting any event not admitted by the topology.
Agents retain full domain-tool autonomy; only coordination operations are constrained.
This design preserves the independence that makes swarm-style MAS attractive while hardening the coordination primitives that are most prone to interleaving-sensitive hazards.

\section{Benchmark}

\subsection{Motivation and design principles}
Prior benchmarks emphasize either collaboration outcomes or formal-model generation quality, but not protocol synthesis for concurrent coordination with shared resources and asynchronous channels~\cite{zhu2025multiagentbench,cheng2026sysmobench}.
Coordination strategies are often benchmark-constrained or task-implied, reducing the need for original protocol design.

We target the complementary setting: given only a natural-language task description, an LLM must synthesize the coordination protocol.
Tasks specify participants, resources, directed channels, and domain tools, but not lock ordering, handshake structure, or control flow.
Ground truth is coordination correctness (deadlock freedom, mutual exclusion, channel drainage), with TLC as the primary evaluator.

Design follows five constraints.
\emph{LLM relevance}: scenarios reflect domains where LLM agents are plausible operators (planning, communication, tool use), spanning software development, research writing, medical consultation, manufacturing, pharmaceutical research, and semiconductor fabrication.
\emph{Non-obvious coordination}: tasks provide no protocol templates or pattern hints.
\emph{Simulatable environments}: all 48 tasks include simulation harnesses; families 12--16 add richer domain tools and configurable fault injection.
\emph{Formalizable properties}: correctness is encoded as TLC-checkable safety invariants, not domain-semantic quality.
\emph{Multi-dimensional complexity}: difficulty varies jointly across agents, resources, channel topology, and constraint density.

Complexity is therefore not monotone in agent count; e.g., a 5-agent task with 12 channels can be harder than a 7-agent lock-only task because channel interleavings grow combinatorially.

\subsection{Task matrix}
The benchmark contains 48 tasks drawn from 16 scenario families, each instantiated at Easy, Medium, and Hard tiers.
Table~\ref{tab:scenario-summary} summarizes the families; the full 48-row inventory appears in Appendix~\ref{app:task-inventory}.
Higher tiers generally add agents (most Easy tasks have 3, Hard up to 7), shared resources, and channels; topologies range from 6 to 33 total elements.
Scenario~16 corresponds to a continuous integration/continuous delivery (CI/CD) pipeline setting.

The table reveals several structural patterns.
Lock-only tasks (scenarios 1, 2, 9) have zero channels and exhibit predictable state-space growth dominated by agent count; these test pure mutual-exclusion reasoning.
Channel-heavy tasks (scenarios 3, 4, 6, 7, with up to 24 channels at Hard tier) exhibit much larger and more variable state spaces because message ordering introduces factorial interleavings; these test the ability to design deadlock-free handshake protocols.
Tasks with counters (5H, 10H, 11H, 14H, 15H, 16H) model consumable shared resources (e.g., biological samples, wafer slots, build server capacity) and introduce a third axis of nondeterminism.

\begin{table}[t]
\centering
\caption{Scenario summary (16 families $\times$ 3 tiers = 48 tasks). Ranges span E/M/H. $R$ = total repairs across tiers.}
\label{tab:scenario-summary}
\small
\begin{tabular}{@{}rlccccr@{}}
\toprule
\# & Scenario & $|A|$ & $|L|$ & $|C|$ & $R$ & $S$ range \\
\midrule
1 & Shared Codebase Dev & 3--7 & 3--7 & 0 & 0 & 525--2.1\,M \\
2 & Research Paper & 3--7 & 3--7 & 0 & 0 & 525--2.1\,M \\
3 & Report / Survey & 3--7 & 2--6 & 4--17 & 3 & 434--7.7\,M \\
4 & Software Pipelines & 3--7 & 2 & 6--21 & 1 & 106--887\,K \\
5 & Medical Consult. & 3--7 & 1--4 & 6--13 & 1 & 118--21\,K \\
6 & Code Integration & 3--7 & 1--4 & 4--19 & 3 & 1.5\,K--6.1\,M \\
7 & Document Authoring & 3--7 & 1--2 & 4--24 & 1 & 434--259\,K \\
8 & Fullstack Dev & 3--7 & 2--5 & 3--20 & 1 & 122--12\,K \\
9 & Dining Philosophers & 3--7 & 3--7 & 0 & 0 & 336--867\,K \\
10 & Parallel Build & 3--7 & 2--4 & 3--11 & 6 & 254--2.1\,M \\
11 & Flexible Mfg. & 3--7 & 2--4 & 2--16 & 3 & 355--775\,K \\
12 & Collab.\ Kitchen & 3 & 2--3 & 3 & 0 & 199--3.7\,K \\
13 & Pharma Lab & 3--5 & 2--4 & 3--7 & 1 & 54--11\,K \\
14 & Drug Discovery & 4--7 & 3--4 & 5--12 & 2 & 909--56\,K \\
15 & Semiconductor Fab & 6 & 5 & 11--14 & 2 & 18\,K--56\,K \\
16 & CI/CD Pipeline & 3--7 & 2--6 & 4--12 & 5 & 30--45\,K \\
\bottomrule
\end{tabular}
\end{table}

\subsection{Task examples}
\paragraph{Task 10E (Parallel Build, Easy).}
Two builders compile modules in parallel; both depend on a shared core library that can only be compiled against by one builder at a time.
Builder~A also maintains a shared type-definitions file that Builder~B depends on.
An integrator links the final artifacts after both builders finish.
The topology has 8 elements (3 agents, 2 locks, 3 channels); correctness requires mutual exclusion on both locks, no deadlock, and all channels drained on termination.
This task is representative of the Easy tier; the lock ordering is straightforward and the channel structure is acyclic.

\paragraph{Task 9H (Dining Philosophers, Hard).}
Seven philosophers alternate between thinking and eating; each must hold both adjacent forks (locks) to eat.
The description specifies seven fork locks and no channels.
Despite having only 14 topology elements, the state space reaches 867\,K distinct states because each philosopher's think/eat cycle creates $O(2^n)$ reachable states.
This task serves as a useful calibration point; it is well-studied in the concurrency literature~\cite{dijkstra1971hierarchical}, and the LLM correctly produces an ordered-acquisition strategy on the first attempt, confirming that pure lock-ordering coordination is within current LLM capability even at scale.

\subsection{Benchmark objectives}
The benchmark is designed to answer three pragmatic questions.
First, can an LLM design a safe coordination protocol from the task description alone?
This is measured by the first-attempt TLC pass rate.
Second, how does protocol design degrade with complexity, and which coordination hazards dominate?
This is measured by the repair rate and failure taxonomy across difficulty tiers.
Third, does counterexample-driven repair converge within a small iteration budget?
This is measured by the maximum repair depth across all tasks.
Together, these three axes characterize whether the generate/check/repair paradigm is viable for coordination protocol design.
No existing multi-agent benchmark targets the specific combination of concurrent coordination, shared resources, and asynchronous channels that TraceFix addresses (Section 5). We encourage future work to evaluate the pipeline on independently designed coordination tasks; the artifact release includes the full pipeline and can be applied to new task descriptions without modification.

\section{Evaluation}

The evaluation asks four questions, each addressed by a dedicated experiment.
(1)~Can an LLM design a model-checkable coordination protocol, and if the initial attempt fails, does counterexample-driven repair converge?
(2)~What are the dominant root causes of verification failure, and how do they distribute across difficulty tiers?
(3)~How does the state space grow with topology complexity, and does bounded model checking remain tractable?
(4)~Does the verified protocol improve runtime task completion compared to prompt-only and chat-only baselines, and how sensitive is that advantage to model capability and fault injection?

Questions 1--3 are evaluated on the full 48-task benchmark using Claude Opus 4.6 for protocol design.
Question~4 is evaluated on all 48 tasks (each has a simulation harness; families 12--16 have richer domain tools and fault injection) under four runtime architectures, two LLM tiers (gpt-5-mini and gpt-5-nano), three failure-injection levels, and three repetitions per configuration, yielding $48 \times 4 \times 3 \times 3 \times 2 = 3{,}456$ runs.
TLC runs safety-only verification with 12 workers, 4096\,MB heap, breadth-first search, and a default per-channel bound of $B_c = 3$.

\subsection{Protocol design and repair convergence}

\begin{figure}[t]
  \centering
  \includegraphics[width=\columnwidth]{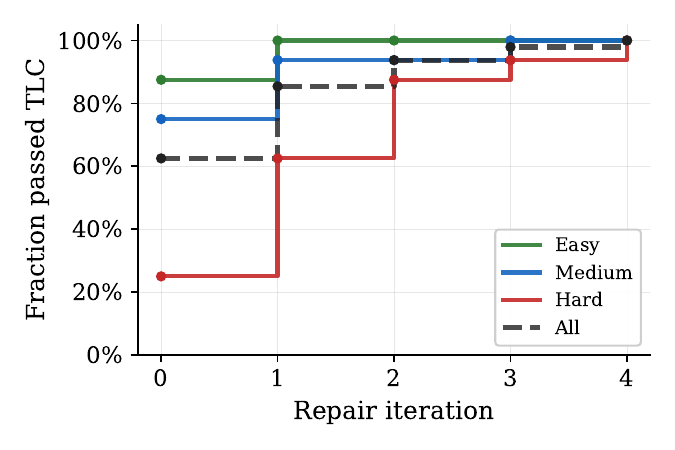}
  \caption{Cumulative TLC verification pass rate as a function of repair iteration, stratified by difficulty.
  62.5\% of tasks require zero repairs; the curve reaches 100\% by iteration~4, with Easy and Medium converging by iterations 1--3.
  \textbf{Takeaway;} the generate/check/repair loop is both necessary (37.5\% fail initially) and fast-converging; counterexample traces provide sufficient signal to close all remaining violations within a small iteration budget.}
  \label{fig:convergence}
\end{figure}

All 48 tasks reach TLC verification (100\% final pass rate).
30 of 48 (62.5\%) pass on the first attempt with zero repairs; the remaining 18 converge after 1--4 repair iterations (Figure~\ref{fig:convergence}).
Easy tasks pass first-attempt in 14 of 16 cases (87.5\%), Medium in 12 of 16 (75.0\%), and Hard in only 4 of 16 (25.0\%).
Hard tasks are 6$\times$ more likely to require repair than Easy tasks and account for 72\% of all repair attempts, confirming that the difficulty tiers capture a genuine gradient in coordination complexity.

The repair loop is essential; more than one third of tasks fail initial verification.
At the same time, convergence is fast; no task requires more than four iterations, and the majority of repaired tasks converge after a single repair.
The higher first-attempt rate for Easy tasks likely reflects simpler topology structure (fewer channels, no counters, straightforward lock ordering), which provides enough structural constraint to guide LLM-generated coordination logic even without counterexample feedback.

The 29 total repair attempts reveal a concentrated failure taxonomy.
Deadlocks dominate (20/29, 69.0\%), with two recurring mechanistic patterns.
First, \texttt{either/or} constructs that commit to a branch before checking channel availability (7~cases); PlusCal's \texttt{either/or} is atomic at the label level, so an agent that places separate \texttt{receive} operations at different labels within an \texttt{either/or} block forces TLC to commit to one branch before the channel state is known, creating an unreachable receive on the unchosen path.
The repair is mechanical; moving receives into the same label as the \texttt{either} keyword so that branch selection and message availability are evaluated atomically.
Second, hub agents that terminate before satellite agents finish retry loops (6~cases); the coordinating agent exits after processing $N$ messages, but a failed tool call causes a satellite to re-send, producing message $N{+}1$ that the hub will never consume.

TLC state-space timeouts account for 5 attempts (17.2\%), all on Hard tasks with retry loops and nondeterministic branching.
These are resolved by reducing $B_c$ or simplifying retry semantics in the model while preserving them in the runtime simulation.
PlusCal syntax errors account for 3 (10.3\%) and an undrained channel for 1 (3.4\%).

\begin{figure}[t]
  \centering
  \includegraphics[width=\columnwidth]{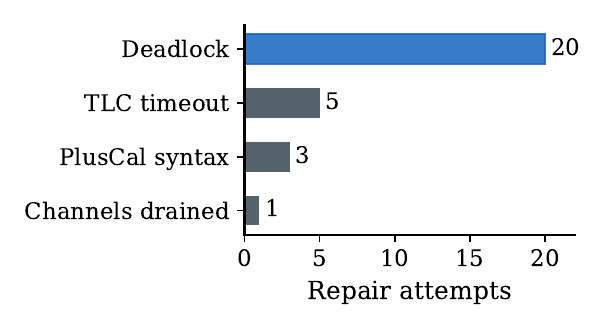}
  \caption{Root-cause distribution of the 29 repair attempts across all 48 tasks.
  Deadlocks dominate (20 attempts, 69.0\%), followed by TLC state-space timeouts (5), PlusCal syntax errors (3), and undrained channels (1).
  The dominant failures are structural coordination hazards (nondeterministic receives, early hub termination) rather than fundamental protocol design flaws.}
  \label{fig:repair-taxonomy}
\end{figure}

\begin{figure}[t]
  \centering
  \includegraphics[width=\columnwidth]{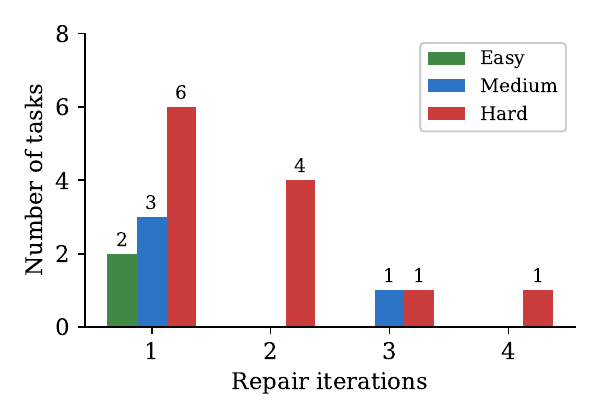}
  \caption{Distribution of repair-requiring tasks (18 of 48) by repair iteration count and difficulty tier.
  Most tasks need a single repair; only one task (10H, Parallel Build Hard) requires 4 iterations.
  Hard tasks dominate (12 of 18) but 2 Easy and 4 Medium tasks also need repair, indicating that topology structure matters alongside the coarse difficulty label.}
  \label{fig:repairs-per-task}
\end{figure}

An important structural observation is that lock-only tasks (scenarios 1, 2, 9) pass first-attempt in all 9 cases.
Alphabetical lock ordering is a well-known pattern that LLMs reproduce reliably.
Every repair-requiring failure arises from the interaction of message-passing channels with retry loops, conditional branching, or multi-step handshakes, confirming that channel semantics, not lock semantics, are the primary source of verification complexity in LLM-generated protocols (Figures~\ref{fig:repair-taxonomy}--\ref{fig:repairs-per-task}).

\subsection{State-space tractability}

For each verified protocol, we record the number of distinct states explored by TLC and the wall-clock verification time.
We define \emph{topology size} as the sum of agents, locks, counters, and channels, providing a single scalar proxy for protocol complexity.

\begin{figure}[t]
  \centering
  \includegraphics[width=\columnwidth]{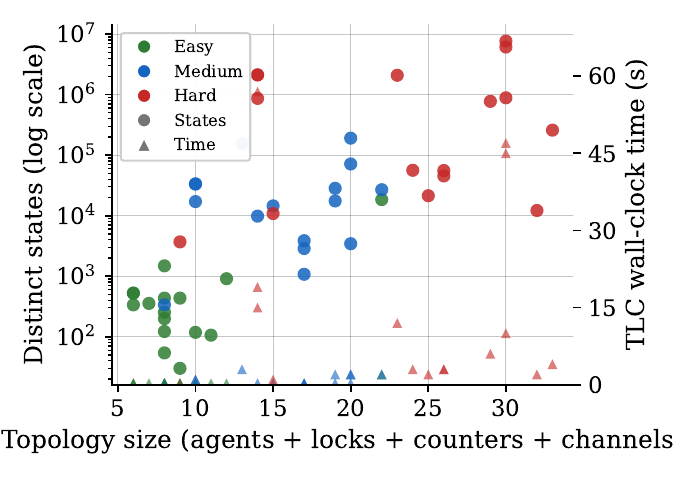}
  \caption{TLC distinct states (circles, left axis, log scale) and wall-clock verification time (triangles, right axis, linear) vs.\ topology size, colored by difficulty.
  State spaces span six orders of magnitude (30 to 7.7\,M) yet verification time remains flat (median $<$1\,s, all under 60\,s).
  \textbf{Takeaway;} PlusCal protocols generated by TraceFix remain within the tractable regime of bounded model checking even for the largest topologies in the benchmark.}
  \label{fig:state-tractability}
\end{figure}

Distinct states range from 30 (task 16E, a 3-agent CI/CD pipeline) to 7.7\,M (task 3H, a 7-agent survey with 17 channels and 6 locks).
Figure~\ref{fig:state-tractability} plots both distinct states and verification time against topology size.
The data reveal three regimes.
Easy tasks cluster tightly below $10^4$ states at topology sizes 6--22, with low variance; the coordination structures are simple enough that agent count dominates state-space size.
Medium tasks span $10^2$--$10^{6}$, reflecting the diversity of channel topologies; manufacturing and build scenarios with counters and nondeterministic retry loops generate disproportionately large state spaces relative to topology size.
Hard tasks exhibit the greatest variance, ranging from $\sim$3.7\,K to 7.7\,M at topology sizes 9--33; the low end reflects compact scenarios (e.g., 12H with 3 agents and 3 channels) while the high end reflects dense channel topologies.

The critical observation is that \emph{verification time does not track state-space size}.
Despite a six-order-of-magnitude range in distinct states, TLC completes in under 60\,s for every task (median $<$1\,s).
The three tasks exceeding 30\,s (1H at 57\,s, 3H at 45\,s, 6H at 47\,s) all involve 7 agents with dense coordination topologies (7 locks for 1H; 17 and 19 channels for 3H and 6H), confirming that the combination of agent count and coordination density, rather than state-space magnitude, drives verification cost.
This decoupling between state-space magnitude and wall-clock time is characteristic of BFS model checking with efficient state hashing~\cite{clarke1999modelchecking}; it means that the protocols TraceFix produces remain within the tractable regime even when the underlying state space is large.

\subsection{Runtime architecture comparison}

The verification pipeline produces a protocol that is correct under bounded assumptions.
Whether that protocol improves actual multi-agent task execution depends on how it is operationalized at runtime.
We evaluate four architectures that represent a spectrum from full enforcement to no coordination infrastructure.

The four conditions are;
\textbf{Topology-monitored}, verified protocol plus coordination tools; a monitor validates each operation against the topology (channels, resources, labels) and rejects violations in real time; step-order discipline is carried by prompts;
\textbf{Mediator-enforced}, a mediator structurally prevents coordination violations by managing all lock and channel operations on behalf of agents; violations are impossible by construction;
\textbf{Prompt-only}, identical tools and protocol-derived prompts but the monitor validates only tool existence, not topology; isolates the contribution of topology-aware monitoring;
\textbf{Chat-only}, agents communicate via a shared chat room with no coordination tools, locks, channels, or state machine.

\paragraph{Metrics.}
\emph{Avg Sim~\%}: average fulfillment of task-specific objectives across all simulation checkpoints (each checkpoint is an objective; partial progress is rewarded).
\emph{Sim~100\%}: fraction of runs where every objective is met.
\emph{DL/LL~\%}: fraction of runs ending in deadlock or livelock.
\emph{Contention~\%}: fraction of tool calls that encounter a resource conflict (concurrent access to an exclusive resource).

\begin{figure}[t]
  \centering
  \includegraphics[width=\columnwidth]{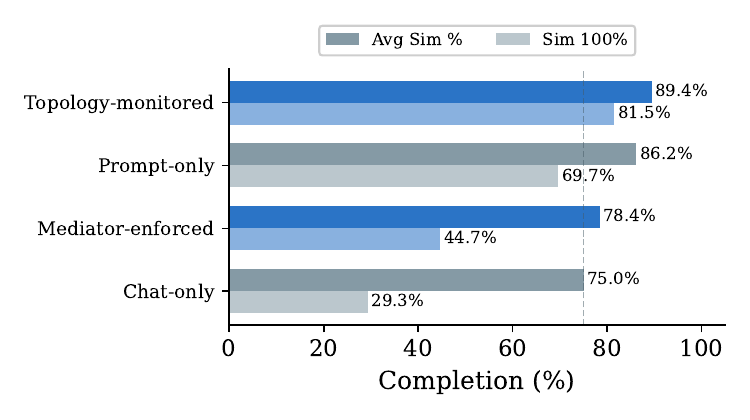}
  \caption{Average and full simulation completion for each runtime architecture (gpt-5-mini, 432 runs per condition).
  Topology-monitored achieves 89.4\% average and 81.5\% full completion; Chat-only achieves only 29.3\% full completion with 61\% resource conflicts.
  \textbf{Takeaway;} verified protocol with topology-aware monitoring yields the best combination of safety and completion; Mediator-enforced prevents violations but cannot adapt to domain-tool failures, depressing completion to 44.7\%.}
  \label{fig:runtime-comparison}
\end{figure}

\paragraph{Results (gpt-5-mini).}
Table~\ref{tab:runtime-mini} and Figure~\ref{fig:runtime-comparison} summarize results across all 48 tasks.
Topology-monitored leads on both completion metrics (89.4\% Avg Sim, 81.5\% Sim~100\%) and wall-clock efficiency (93\,s average, 32\,s faster than the next-best condition; Appendix Table~\ref{tab:safety-metrics}).
The combination of agent autonomy and topology-aware monitoring produces the best outcome; agents can adapt to domain-tool failures (which rigid enforcement cannot handle) while the monitor catches coordination deviations (which Prompt-only agents frequently commit).

Prompt-only achieves the second-highest completion (86.2\% / 69.7\%), confirming that protocol-derived prompts alone provide substantial value.
However, the gap between Prompt-only and Topology-monitored is largest on the strict metric (69.7\% vs.\ 81.5\%, a 12\,pp difference), and Prompt-only's deadlock rate is 2.4$\times$ higher (21.1\% vs.\ 8.8\%).
This demonstrates that prompts guide agents toward the correct protocol but are insufficient to prevent coordination failures under rare interleavings; topology monitoring closes the gap.

Mediator-enforced achieves the lowest deadlock rate (4.5\%) and lowest resource contention (1.9\%), consistent with its design; the mediator structurally prevents coordination violations.
However, it also achieves the lowest completion among non-chat runtimes (78.4\% / 44.7\%).
The mechanism is instructive; when a domain tool call fails unexpectedly, the mediator has no adaptation path, leaving unfinished objectives.
The 34\,pp gap between Avg Sim and Sim~100\% confirms that Mediator-enforced runs frequently achieve high partial completion but stall on the last few objectives.
This reveals a fundamental tension; enforcement maximizes safety but sacrifices the flexibility that agents need to recover from domain-level surprises.

Chat-only is worst on every metric.
Without coordination tools, 61.2\% of tool calls encounter resource conflicts, 32.6\% of runs end in deadlock or livelock, and only 29.3\% complete all objectives.
Agents spend roughly 4$\times$ more tool-call steps than Mediator-enforced (203 vs.\ 53; Appendix Table~\ref{tab:safety-metrics}) attempting to self-organize coordination through natural-language negotiation, largely unsuccessfully.

\begin{table}[t]
\centering
\caption{Runtime comparison (gpt-5-mini, 48 tasks, 432 runs per runtime). DL/LL = deadlock or livelock rate. Contention\,\% = fraction of tool calls where an agent attempts to access an exclusive resource already held by another agent; Chat-only's 61.2\% contention indicates that over 3 in 5 tool calls fail due to resource conflicts.}
\label{tab:runtime-mini}
\small
\begin{tabular}{@{}lcccc@{}}
\toprule
Architecture & Avg Sim\,\% & Sim 100\,\% & DL/LL\,\% & Contention\,\% \\
\midrule
Topology-monitored  & \textbf{89.4} & \textbf{81.5} &  8.8 &  3.2 \\
Prompt-only         & 86.2 & 69.7 & 21.1 &  7.3 \\
Mediator-enforced   & 78.4 & 44.7 &  \textbf{4.5} &  \textbf{1.9} \\
Chat-only           & 75.0 & 29.3 & 32.6 & 61.2 \\
\bottomrule
\end{tabular}
\end{table}

\begin{figure}[t]
  \centering
  \includegraphics[width=\columnwidth]{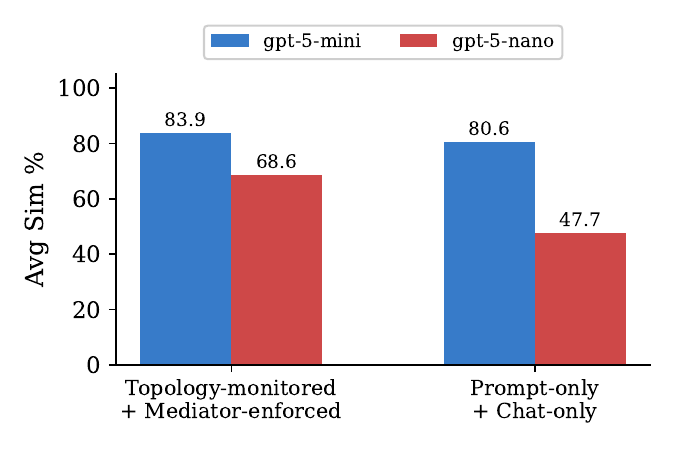}
  \caption{Average simulation completion for Topology-monitored + Mediator-enforced (pooled) vs.\ Prompt-only + Chat-only (pooled) under gpt-5-mini and gpt-5-nano.
  Verified-protocol runtimes degrade by $\sim$15\,pp; prompt-only and chat-only by $\sim$33\,pp, a 2$\times$ gap.
  \textbf{Takeaway;} the verified protocol acts as a capability buffer, absorbing roughly half the performance loss from using a weaker model.}
  \label{fig:verification-buffer}
\end{figure}

\paragraph{Model degradation.}
Switching from gpt-5-mini to gpt-5-nano reveals a strong verification buffer.
Topology-monitored and Mediator-enforced lose about 15\,pp in Avg Sim, while Prompt-only and Chat-only lose about 30--36\,pp (Figure~\ref{fig:verification-buffer}).
The ranking swap between Prompt-only and Mediator-enforced across model tiers indicates that enforcement becomes more valuable as model capability drops.
Chat-only degrades most severely and is not viable at high difficulty under faults.

\begin{figure}[t]
  \centering
  \includegraphics[width=\columnwidth]{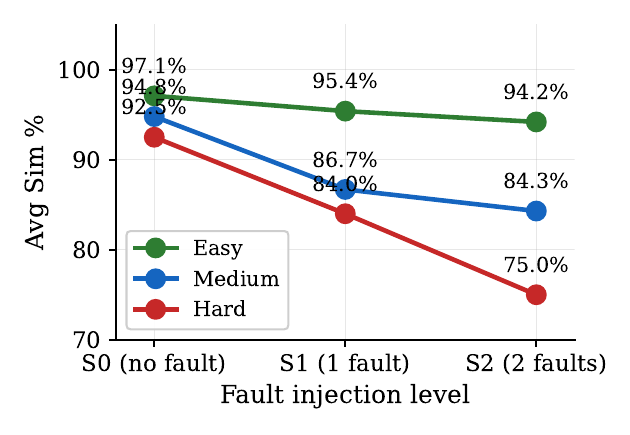}
  \caption{Avg Sim~\% for Topology-monitored (gpt-5-mini) across difficulty tiers and failure-injection levels.
  Performance degrades gracefully from top-left (Easy, no faults; 97.1\%) to bottom-right (Hard, 2 faults; 75.0\%).
  The verified protocol provides a structured recovery framework that sustains high completion even under increasing fault injection.}
  \label{fig:difficulty-scenario}
\end{figure}

\paragraph{Fault tolerance and safety.}
All runtimes degrade from S0 (no injected faults) through S1 (one injected fault) to S2 (two injected faults), but rankings remain stable across difficulty and fault settings.
Topology-monitored drops from 94.8\% to 84.5\% Avg Sim on gpt-5-mini, showing graceful degradation under injected failures.
The verified protocol's explicit branch structure supports retries and fallback while preserving coordination constraints (Figure~\ref{fig:difficulty-scenario}).

Lock violations show the safety gap clearly (Appendix Table~\ref{tab:safety-metrics}).
Topology-monitored and Mediator-enforced record zero lock violations across both models, while Chat-only accumulates 879 (gpt-5-mini) to 1{,}540 (gpt-5-nano).
Topology-monitored also rejects 1{,}354 invalid coordination attempts on gpt-5-mini, rising to 2{,}350 on gpt-5-nano (+74\%), confirming that runtime monitoring becomes more valuable as model capability decreases.

\subsection{Ablation; verified vs.\ unverified protocols}

The benchmark-level comparison above isolates runtime architecture.
We also run a paired ablation that isolates protocol quality by holding runtime fixed and varying only whether the protocol passes TLC verification.
Across 15 paired tasks under the Topology-monitored runtime (270 runs total), verified protocols reduce DL/LL from 31.1\% to 14.1\% and increase average completion from 72.6\% to 82.1\%.
The separation is smallest in S0 and largest in S1/S2, confirming that counterexample-guided repair matters most under retries and adverse interleavings.
The effect is strongest on deadlock-labeled tasks, while timeout-labeled tasks are mixed, consistent with the semantic-drift discussion in Section~\ref{sec:limitations}.
Appendix~\ref{app:ablation-6h} gives a concrete protocol-level case study.

\section{Related Work}

\paragraph{Protocol completion and TLA+ synthesis.}
Counterexample-guided loops for distributed protocol completion alternate candidate synthesis with model checking. Alur et al.~\cite{alur2015completion} formalize this counterexample-guided inductive synthesis (CEGIS) approach with SMT completion and symmetry constraints; Scythe~\cite{egolf2024scythe} extends it to TLA+, synthesizing protocol sketches from counterexamples. These target classical message-passing protocols; TraceFix adapts the loop for LLM-synthesized agent coordination (topology IR + PlusCal) and adds runtime enforcement. TLA+ has proven practical for real systems~\cite{newcombe2015amazon}; TraceFix brings that discipline to LLM-generated protocols.

\paragraph{Multiparty session types and runtime enforcement.}
Multiparty Session Types (MPST)~\cite{honda2008mpst} provides a type-theoretic discipline for global multiparty protocols with deadlock-freedom guarantees; monitored semantics~\cite{bocchi2017monitoring} support decentralized runtime enforcement. MPST assumes manually authored protocols; TraceFix targets LLM-generated protocols, uses TLC counterexamples for repair~\cite{lamport-tla}, and deploys a topology monitor that rejects out-of-protocol operations.

\paragraph{LLM orchestration and failure taxonomies.}
Mainstream LLM multi-agent frameworks (AutoGen~\cite{wu2023autogen}, LangGraph~\cite{langgraph}, role-based systems~\cite{crewai}) encode coordination in prompts and workflow graphs without formal semantics or exhaustive correctness checks. Empirical taxonomies such as MAST~\cite{cemri2025why} document that many failures stem from specification and orchestration gaps (missed handshakes, premature termination), rather than single-agent reasoning errors. TraceFix targets the subset expressible as coordination safety properties. Substrates such as Google Agent Development Kit (ADK)~\cite{google-adk} provide concurrent execution with shared state, creating the coordination hazards our verification-first layer addresses. The compound AI systems paradigm~\cite{chen2025optimizing} emphasizes principled composition of system components, including optimizers like DSPy~\cite{khattab2024dspy} that tune individual module quality. TraceFix is complementary: DSPy optimizes what each agent produces; TraceFix verifies how agents coordinate. A pipeline could use DSPy to optimize per-agent prompts and TraceFix to verify the coordination protocol that connects them.

\paragraph{Verification-guided generation and repair.}
Generate/check/repair loops appear in VERGE~\cite{singh2026verge} (LLMs with SMT), Clover~\cite{sun2024clover} (consistency-checking of code and annotations), and SymbolicAI~\cite{dinu2024symbolicai} (solver-in-the-loop); these target functional correctness of single artifacts. TraceFix targets coordination correctness of concurrent multi-agent protocols. ModelWisdom~\cite{chen2026modelwisdom} and TLA+ explorers~\cite{tla-web} improve counterexample usability; our repair loop targets generation-time editing of PlusCal source. Ramani et al.~\cite{ramani2025bridging} translate plans to Kripke structures~\cite{kripke1963semantical} and linear temporal logic (LTL)~\cite{4567924}; our artifact is an executable concurrent program with deadlock-freedom and channel-discipline properties.

\paragraph{Runtime enforcement and benchmarks.}
Classical runtime verification (RV) has long studied monitors that reject invalid events~\cite{leucker2009rv}; our topology monitor applies this pattern to coordination operations. AgentGuard~\cite{koohestani2025agentguard} and Pro2Guard~\cite{wang2025pro2guard} use probabilistic model checking over learned abstractions; ShieldAgent~\cite{chen2025shieldagent} and RvLLM~\cite{zhang2025rvllm} enforce safety policies or validate outputs. Our monitor is a lightweight protocol-conformance layer for locks and channels. MultiAgentBench~\cite{zhu2025multiagentbench} evaluates collaboration; SysMoBench~\cite{cheng2026sysmobench} evaluates TLA+ model generation. Our benchmark makes coordination correctness the primary objective. Position papers~\cite{zhang2024fusion,ferrari2025requirements} argue for tighter LLM/formal-methods integration; TraceFix is a concrete instantiation for coordination.

\section{Limitations and Scope}
\label{sec:limitations}

We verify coordination safety under bounded assumptions: deadlock freedom (TLC's \texttt{NoDeadlock}), mutual exclusion, no orphan locks, channel drainage.
Liveness is not checked because LLM agent protocols routinely contain nondeterministic loops that are valid business logic (e.g., a reviewer may reject indefinitely); requiring guaranteed termination would reject correct protocols.
Deadlock freedom already covers the most dangerous coordination failures---states where no agent can act---and the bounded state space ensures that this check is exhaustive within the modeled bounds.
Task semantics and output quality are likewise out of scope.

TLC explores exhaustively within bounds (finite agents, bounded counters, bounded queues); outside those bounds the guarantee is incomplete.

TLC verifies the full PlusCal step sequence; the runtime enforces only topology conformance and mutual exclusion, not step ordering, lock order, handshake discipline, or receive patterns.
Agents can still reintroduce deadlock through step-order deviations that topology checks cannot catch.
The paired ablation shows that repaired prompts reduce this risk, but a residual gap remains because runtime enforcement is topology-level rather than label-level.

Timeout-driven repair can also induce semantic drift on the hardest tasks.
When TLC repeatedly times out, repairs that simplify branching and retries can reduce checker load but also weaken fidelity to task-level coordination semantics.
The paired ablation shows that this effect concentrates in timeout-labeled tasks; one task shows no pass/failed separation and another shows only moderate separation, unlike the clear gains on deadlock-labeled tasks.
The resulting guarantee remains valid for the repaired model, but coverage narrows when simplification removes coordination behavior that the original task intended to preserve.

The pipeline uses a single LLM (Claude Opus 4.6) for protocol design; runtime comparison uses two OpenAI models. Generalization to other protocol-design models remains open.

\section{Conclusion}

TraceFix applies counterexample-driven model checking to LLM-synthesized coordination protocols.
Empirical results support four claims.
(1)~The generate/check/repair loop is necessary and sufficient; 37.5\% of tasks fail initial verification yet all 48 converge within four iterations, with structural hazards (nondeterministic receives, early hub termination) dominating over fundamental design flaws.
(2)~Bounded model checking remains tractable; state spaces span six orders of magnitude yet verification completes in under 60\,s for every task.
(3)~The verified protocol yields measurable runtime benefit; Topology-monitored achieves the highest completion (89.4\% avg, 81.5\% full) and runtimes using it degrade at roughly half the rate of Prompt-only and Chat-only when model capability is reduced.
(4)~A paired ablation confirms that counterexample-guided repair removes concrete coordination hazards whose runtime impact is largest under fault injection.
Open directions include closing the verification-to-enforcement gap (Section~\ref{sec:limitations}), reducing semantic drift in timeout-driven repair, and extending ablations that isolate monitoring and step-order enforcement independently.

\begin{acks}
This work was supported by the National Science Foundation (NSF) as part of the Center for Smart Streetscapes, under NSF Cooperative Agreement EEC-2133516.
\end{acks}

\bibliographystyle{ACM-Reference-Format}
\bibliography{refs}

\appendix
\section{Full Task Inventory}
\label{app:task-inventory}

Table~\ref{tab:full-inventory} lists all 48 tasks. $|A|$ = agents, $|L|$ = locks, $|K|$ = counters, $|C|$ = channels, $R$ = repairs, $S$ = distinct states.
This appendix table is intended as the structural backbone for the benchmark discussion in the main text.
Rather than repeating per-scenario descriptions, we expose the full design surface in one place so readers can trace how coordination complexity changes across tiers and domains.
The key pattern to read for is not only scale (more agents/channels), but interaction type: lock-only tasks cluster with zero repairs, while channel- and counter-heavy tasks account for most repair activity and larger verification search spaces.

\begin{table*}[t]
\centering
\caption{Full task inventory (48 tasks, 16 scenarios $\times$ 3 tiers). Legend in section text.}
\label{tab:full-inventory}
\small
\setlength{\tabcolsep}{4pt}
\begin{tabular}{@{}lcccccr|@{\quad}lcccccr@{}}
\toprule
\multicolumn{7}{c}{Scenarios 1--8} & \multicolumn{7}{c}{Scenarios 9--16} \\
\cmidrule(lr){1-7} \cmidrule(lr){8-14}
Task & $|A|$ & $|L|$ & $|K|$ & $|C|$ & $R$ & $S$ &
Task & $|A|$ & $|L|$ & $|K|$ & $|C|$ & $R$ & $S$ \\
\midrule
1E  & 3 & 3 & 0 &  0 & 0 & 525   & 9E  & 3 & 3 & 0 &  0 & 0 & 336   \\
1M  & 5 & 5 & 0 &  0 & 0 & 33K  & 9M  & 5 & 5 & 0 &  0 & 0 & 17K  \\
1H  & 7 & 7 & 0 &  0 & 0 & 2.1M & 9H  & 7 & 7 & 0 &  0 & 0 & 867K \\
2E & 3 & 3 & 0 &  0 & 0 & 525   & 10E & 3 & 2 & 0 &  3 & 1 & 254   \\
2M  & 5 & 5 & 0 &  0 & 0 & 33K  & 10M & 5 & 3 & 0 &  7 & 1 & 15K  \\
2H  & 7 & 7 & 0 &  0 & 0 & 2.1M & 10H & 7 & 4 & 1 & 11 & 4 & 2.1M \\
3E  & 3 & 2 & 0 &  4 & 0 & 434   & 11E & 3 & 2 & 0 &  2 & 0 & 355   \\
3M  & 5 & 2 & 0 & 13 & 1 & 72K  & 11M & 5 & 3 & 1 &  4 & 0 & 154K \\
3H  & 7 & 6 & 0 & 17 & 2 & 7.7M & 11H & 7 & 4 & 2 & 16 & 3 & 775K \\
4E  & 3 & 2 & 0 &  6 & 0 & 106   & 12E & 3 & 2 & 0 &  3 & 0 & 199   \\
4M  & 5 & 2 & 0 & 12 & 0 & 18K  & 12M & 3 & 2 & 0 &  3 & 0 & 337   \\
4H  & 7 & 2 & 0 & 21 & 1 & 887K & 12H & 3 & 3 & 0 &  3 & 0 & 3.7K  \\
5E  & 3 & 1 & 0 &  6 & 0 & 118   & 13E & 3 & 2 & 0 &  3 & 0 & 54    \\
5M  & 5 & 3 & 0 & 12 & 0 & 3.4K & 13M & 4 & 3 & 0 &  7 & 0 & 9.8K  \\
5H  & 7 & 4 & 1 & 13 & 1 & 21K  & 13H & 5 & 4 & 0 &  6 & 1 & 11K   \\
6E  & 3 & 1 & 0 &  4 & 0 & 1.5K & 14E & 4 & 3 & 0 &  5 & 0 & 909   \\
6M  & 5 & 3 & 0 & 12 & 1 & 191K & 14M & 5 & 3 & 0 &  9 & 0 & 2.9K  \\
6H  & 7 & 4 & 0 & 19 & 2 & 6.1M & 14H & 7 & 4 & 1 & 12 & 2 & 56K   \\
7E  & 3 & 1 & 0 &  4 & 0 & 434   & 15E & 6 & 5 & 0 & 11 & 1 & 18K   \\
7M  & 5 & 2 & 0 & 12 & 0 & 28K  & 15M & 6 & 5 & 0 & 11 & 0 & 27K   \\
7H  & 7 & 2 & 0 & 24 & 1 & 259K & 15H & 6 & 5 & 1 & 14 & 1 & 56K   \\
8E  & 3 & 2 & 0 &  3 & 0 & 122   & 16E & 3 & 2 & 0 &  4 & 0 & 30    \\
8M  & 5 & 3 & 0 &  9 & 0 & 1.1K & 16M & 5 & 4 & 0 &  8 & 3 & 3.9K  \\
8H  & 7 & 5 & 0 & 20 & 1 & 12K  & 16H & 7 & 6 & 1 & 12 & 2 & 45K   \\
\bottomrule
\end{tabular}
\end{table*}

\section{Primer; TLA+, PlusCal, and TLC}
\label{app:primer}

This paper uses TLA+ as the formal representation of coordination protocols.
At a high level, a TLA+ specification defines a set of system states and transition rules that describe how the system may evolve.
Safety properties are then stated as invariants over those states (for example, ``no two agents hold the same lock'' or ``all channels are drained when all agents terminate'').

We author protocols in PlusCal, an algorithm-like language designed to be easier to read and write than raw TLA+.
PlusCal code is automatically translated to TLA+ (\texttt{pcal.trans}), preserving control flow and synchronization structure in a checker-friendly form.
This lets us keep protocol logic close to standard pseudocode while still obtaining formal analysis.

TLC is the TLA+ model checker.
Given a translated specification and bounded parameters, TLC explores reachable executions and checks whether safety properties hold.
If a property fails, TLC produces a concrete counterexample trace (an explicit failing execution), which TraceFix uses as repair signal in the generate/check/repair loop.

\section{End-to-End Example; Task 3H}
\label{app:example-3h}

\begingroup
\setlength{\parskip}{0.5ex plus 0.3ex}
Task~3H (Large Survey Paper Production, Hard) has 7 agents, 6 locks, and 17 channels.
Four researchers each cover a subtopic, a data analyst produces figures and tables, a reviewer provides peer review, and an editor-in-chief assembles the complete paper.
We include this example to make the generate/check/repair loop concrete at trace level.
The task exhibits a structural hazard class common in channel-heavy protocols: a coordination hub that terminates before satellite agents complete their revision loops.

\paragraph{Generated topology.}
The agent produces a topology with seven agents
(\texttt{researcher\_a}--\texttt{d},
\texttt{data\_analyst}, \texttt{reviewer},
\texttt{editor\_\allowbreak in\_\allowbreak chief}),
six locks
(\texttt{section\_a}--\texttt{d},
\texttt{figure\_\allowbreak repository}, \texttt{database}),
and seventeen directed channels with labels such as
\texttt{data\_ready}, \texttt{figures\_done},
\texttt{submit}, \texttt{accept}, \texttt{revise},
and \texttt{section\_approved}.

\paragraph{First TLC run; deadlock.}
\begin{sloppypar}
TLC discovers a deadlock with a 97-step counterexample trace.
The trace reveals that \texttt{data\_analyst} completes its cycle after handling each researcher's initial data request and terminates.
\texttt{researcher\_d}, having received a \texttt{revise} verdict from the reviewer, re-enters its revision loop and re-sends \texttt{data\_ready}---but \texttt{data\_analyst} will never consume it.
\texttt{researcher\_d} then blocks waiting for \texttt{figures\_done}, the reviewer blocks waiting for \texttt{researcher\_d}'s resubmission, and the editor-in-chief blocks waiting for the reviewer's final \texttt{section\_approved}---a hub-blocks-on-terminated-sender cascade that propagates through three downstream agents.
\end{sloppypar}

\paragraph{Iterative repair; verified.}
The repair agent restructures the data analyst with an \texttt{either/or} at the top level to accept data requests from any researcher in any order, with per-researcher completion flags instead of a fixed-count exit, so that revision-triggered re-requests are still consumed.
A second repair addresses a similar early-termination hazard in the reviewer, replacing approval counting with per-researcher done flags and an explicit \texttt{review\_done} signal.
After two repair iterations, TLC explores 7.7\,M distinct states with zero violations; all safety invariants hold, \texttt{NoDeadlock} passes, and all channels drain on termination.
At deployment, the monitor allows \texttt{researcher\_a} to send only on its declared channels (\texttt{ch\_ra\_to\_da}, \texttt{ch\_ra\_to\_rev}) with valid labels and to acquire only \texttt{section\_a\_lock} and \texttt{database}; any out-of-topology operation is rejected.
This progression (97-step failing trace to 7.7\,M-state verified protocol) illustrates the central claim of the paper: counterexamples provide actionable, localized repair targets that improve both formal verification outcomes and runtime-facing protocol instructions.
\endgroup

\section{Ablation Case Study; Task 6H}
\label{app:ablation-6h}

Task 6H (Parallel Feature Development with Staged Integration) is representative of the deadlock class in the ablation.
Under identical runtime settings, the verified protocol sustains full completion while the failed protocol degrades under fault injection.
This section complements the aggregate ablation numbers by showing the mechanism of separation at code level.
The goal is to connect the quantitative gap (higher completion under S1/S2, i.e., one/two injected faults, for verified protocols) to concrete protocol defects that are present before repair and absent after repair.

\begin{table}[h]
\centering
\caption{Task 6H paired ablation summary (Topology-monitored, gpt-5-mini, 3 runs per scenario).}
\label{tab:ablation-6h}
\small
\begin{tabular}{@{}lcccc@{}}
\toprule
Condition & S0 Sim\,\% & S1 Sim\,\% & S2 Sim\,\% & DL/LL \\
\midrule
Pass (TLC verified) & 100.0 & 100.0 & 100.0 & 0/9 \\
Failed (TLC failed) & 100.0 & 79.2 & 70.8 & 0/9 \\
\bottomrule
\end{tabular}
\end{table}

\paragraph{Bug class 1; circular lock dependency.}
The failed protocol allows a developer and a CI runner to hold different locks and wait on each other.
TLC exposes a reachable circular wait where \texttt{REPO} is held by \texttt{DEVELOPER\_A} while \texttt{BUILD\_SERVER} is held by \texttt{CI\_RUNNER}; both processes block on the other lock.
The repair removes cross-lock acquisition between these roles, producing disjoint lock ownership.

\begin{verbatim}
(* failed *)
DEVELOPER_A:
  acquire_lock(REPO); ... acquire_lock(BUILD_SERVER);
CI_RUNNER:
  acquire_lock(BUILD_SERVER); ... acquire_lock(REPO);

(* repaired *)
DEVELOPER_A:
  acquire_lock(REPO); ... release_lock(REPO);
CI_RUNNER:
  acquire_lock(BUILD_SERVER);
  ... release_lock(BUILD_SERVER);
\end{verbatim}

\paragraph{Bug class 2; premature reviewer termination.}
In the failed protocol, the reviewer exits after three approvals, even though developers can re-enter review after CI or staging setbacks.
This leaves later review requests unconsumed.
The repair replaces approval counting with per-developer completion flags and explicit \texttt{review\_done} signals.

\begin{verbatim}
(* failed *)
while (rvCount < 3) {
  receive(review_request);
  send(approved or revise);
};
goto Done;

(* repaired *)
receive_any(ch_a_review, ch_b_review, ch_c_review);
if (msg = "review_done") { mark_done(agent); }
if (a_done /\ b_done /\ c_done) {
  goto Done;
} else {
  goto rev_loop;
};
\end{verbatim}

\paragraph{Prompt propagation.}
These PlusCal defects are compiled into runtime prompts.
In the failed prompt, \texttt{DEVELOPER\_A} is instructed to hold \texttt{REPO} while attempting \texttt{BUILD\_SERVER}, directly encoding the circular lock dependency.
In the repaired prompt, the build server step is removed from developer merge steps, and failure branches explicitly route to retry paths.
The case demonstrates that counterexample-guided repair improves runtime behavior by changing executable coordination instructions, not only by improving verification status.
Taken together with Table~\ref{tab:ablation-6h}, this example supports the interpretation used in the main evaluation: TLC-guided repair does not merely ``clean up'' specifications, it removes concrete coordination hazards that otherwise amplify under fault injection.

\section{Efficiency and Safety Metrics}
\label{app:efficiency-safety}

Table~\ref{tab:safety-metrics} reports per-runtime efficiency and safety metrics aggregated across 432 runs per runtime per model (48 tasks $\times$ 3 failure scenarios $\times$ 3 repetitions).

\begin{table}[h]
\centering
\caption{Efficiency and safety metrics (432 runs per runtime per model). Steps = mean total tool-call steps per run. Duration = mean wall-clock seconds. Lock Viol.\ = total lock violations across all runs. Monitor Rej.\ = total coordination operations rejected by the topology monitor (Topology-monitored only; not applicable to other runtimes).}
\label{tab:safety-metrics}
\small
\setlength{\tabcolsep}{3pt}
\begin{tabular}{@{}llrrrr@{}}
\toprule
Model & Runtime & Steps & Dur.\,(s) & Lock Viol. & Mon.\ Rej. \\
\midrule
\multirow{4}{*}{mini}
 & Topology-mon.  &  62 &  93 &     0 & 1{,}354 \\
 & Mediator-enf.  &  53 & 125 &     0 & --- \\
 & Prompt-only    &  70 & 136 &     8 & --- \\
 & Chat-only      & 203 & 229 &   879 & --- \\
\midrule
\multirow{4}{*}{nano}
 & Topology-mon.  &  71 & 107 &     0 & 2{,}350 \\
 & Mediator-enf.  &  61 & 144 &     0 & --- \\
 & Prompt-only    &  84 & 163 &    42 & --- \\
 & Chat-only      & 264 & 298 & 1{,}540 & --- \\
\bottomrule
\end{tabular}
\end{table}

\section{Bound Sensitivity}
\label{app:bc-sensitivity}

To assess whether the default per-channel bound $B_c=3$ masks safety violations in channel-dense protocols, we re-run TLC on the five most channel-dense Hard tasks at $B_c \in \{3, 5, 7\}$ with a 300\,s wall-clock budget per run.
Table~\ref{tab:bc-sensitivity} reports outcome and elapsed time for each configuration.

\begin{table}[h]
\centering
\caption{Bound sensitivity on the five most channel-dense Hard tasks under a 300\,s budget per run. ``pass'' = TLC explored the full bounded state space without finding a safety violation; ``timeout'' = TLC exceeded the budget. No new violations are surfaced at higher bounds.}
\label{tab:bc-sensitivity}
\small
\setlength{\tabcolsep}{4pt}
\begin{tabular}{@{}lccc@{}}
\toprule
Task & $B_c=3$ & $B_c=5$ & $B_c=7$ \\
\midrule
3H  & pass & pass & pass \\
4H  & pass & pass & pass \\
6H  & pass & pass & pass \\
8H  & pass & pass & pass \\
11H & timeout & timeout & timeout \\
\bottomrule
\end{tabular}
\end{table}

Four of five tasks pass at every bound; 11H times out at all three bounds---a tractability limit consistent with Section~4.1's observation that some Hard tasks require a reduced $B_c$ or simplified retry semantics to fit the checker budget, not a new safety violation.
This is consistent with the property structure discussed in Section~2.5: deadlock, mutual exclusion, orphan locks, and channel drainage depend on the topology of acquire/release/send/receive operations rather than on how many messages coexist in any single channel.
We do not claim that $B_c=3$ is sufficient for arbitrary protocols; a protocol whose correctness depends on explicit queue-fullness logic could in principle exhibit bound-sensitive behavior, in which case raising $B_c$ would be appropriate.

\section{Protocol Topology IR Schema}
\label{app:ir-schema}

The protocol topology IR is validated structurally before any TLC invocation.
The JSON schema below (Draft~7, \texttt{additionalProperties: false} throughout) specifies the three top-level sections (\texttt{agents}, \texttt{resources}, \texttt{channels}).

\begin{lstlisting}[language=]
{
  "title": "TLA+ MAS Coordination Protocol IR v3 (PlusCal)",
  "description": "Intermediate Representation for coordination protocols between independent concurrent agents. Defines agents, shared resources, and message channels. Agent behavior is written as PlusCal process bodies (not JSON states).",
  "type": "object",
  "required": ["agents", "resources", "channels"],
  "additionalProperties": false,

  "properties": {
    "agents": {
      "description": "Independent agent processes. Each runs concurrently as a PlusCal process.",
      "type": "array",
      "minItems": 1,
      "items": {
        "type": "object",
        "required": ["id"],
        "additionalProperties": false,
        "properties": {
          "id": {
            "type": "string",
            "minLength": 1,
            "description": "Unique agent identifier."
          },
          "tools": {
            "description": "Optional: available tools for this agent. Ignored by TLA+ verification, used by execution runtime.",
            "type": "array",
            "items": {"type": "string", "minLength": 1}
          }
        }
      }
    },

    "resources": {
      "description": "Shared resources that agents compete for. Lock = exclusive access, Counter = counting semaphore.",
      "type": "array",
      "items": {
        "type": "object",
        "required": ["id", "type"],
        "additionalProperties": false,
        "properties": {
          "id": {
            "type": "string",
            "minLength": 1
          },
          "type": {
            "type": "string",
            "enum": ["Lock", "Counter"]
          },
          "config": {
            "type": "object",
            "additionalProperties": false,
            "properties": {
              "initial": {
                "type": "integer",
                "minimum": 0,
                "description": "Initial counter value. Required for Counter type."
              }
            }
          }
        }
      }
    },

    "channels": {
      "description": "Unbounded FIFO message channels between agents. Send always succeeds; receive blocks if empty.",
      "type": "array",
      "items": {
        "type": "object",
        "required": ["id", "from", "to", "labels"],
        "additionalProperties": false,
        "properties": {
          "id": {
            "type": "string",
            "minLength": 1
          },
          "from": {
            "description": "Agent(s) allowed to send on this channel.",
            "oneOf": [
              {"type": "string", "minLength": 1},
              {"type": "array", "items": {"type": "string", "minLength": 1}, "minItems": 1}
            ]
          },
          "to": {
            "description": "Agent(s) allowed to receive from this channel.",
            "oneOf": [
              {"type": "string", "minLength": 1},
              {"type": "array", "items": {"type": "string", "minLength": 1}, "minItems": 1}
            ]
          },
          "labels": {
            "description": "Message types that flow through this channel. All send/receive operations on this channel must use one of these labels.",
            "type": "array",
            "items": {"type": "string", "minLength": 1},
            "minItems": 1
          }
        }
      }
    }
  }
}
\end{lstlisting}

Channels are unbounded FIFO message queues; send always succeeds and receive blocks when empty.
Both \texttt{from} and \texttt{to} accept either a single agent ID or an array of agent IDs to support multi-sender/multi-receiver topologies; a channel's \texttt{labels} array enumerates the message types that may flow through it.
Beyond the schema, the validator enforces semantic invariants: every \texttt{from}/\texttt{to} endpoint references a declared agent; channel and resource IDs are unique; at most one channel exists per directed (from, to) pair (with label vocabulary distinguishing message types within that channel); and \texttt{Counter} resources declare a non-negative \texttt{config.initial} value.
A topology failing any check is rejected before scaffolding into PlusCal.

\paragraph{Example IR (Task 14H).}
The verified IR for Task~14H (Drug Discovery Pipeline, Hard) declares 7 agents, 4 \texttt{Lock} resources, 1 \texttt{Counter} (initial value 8), and 12 directed channels with both forward-progress labels (\emph{compound}, \emph{approval}, etc.) and feedback labels (\emph{retry}, \emph{abort}, etc.):

\begin{lstlisting}[language=]
{
  "agents": [
    {"id": "BIOLOGIST"},
    {"id": "CHEMIST"},
    {"id": "CLINICAL_LEAD"},
    {"id": "FORMULATION_SCIENTIST"},
    {"id": "PROJECT_DIRECTOR"},
    {"id": "REGULATORY_SPECIALIST"},
    {"id": "TOXICOLOGIST"}
  ],
  "resources": [
    {"id": "HPLC", "type": "Lock"},
    {"id": "MASS_SPEC", "type": "Lock"},
    {"id": "CELL_LAB", "type": "Lock"},
    {"id": "FORMULATION_SUITE", "type": "Lock"},
    {"id": "BIOLOGICAL_SAMPLES", "type": "Counter", "config": {"initial": 8}}
  ],
  "channels": [
    {
      "id": "ch_chem_to_form",
      "from": "CHEMIST",
      "to": "FORMULATION_SCIENTIST",
      "labels": ["compound", "retry_compound"]
    },
    {
      "id": "ch_form_to_bio",
      "from": "FORMULATION_SCIENTIST",
      "to": "BIOLOGIST",
      "labels": ["formulated", "done"]
    },
    {
      "id": "ch_form_to_tox",
      "from": "FORMULATION_SCIENTIST",
      "to": "TOXICOLOGIST",
      "labels": ["formulated", "done"]
    },
    {
      "id": "ch_bio_to_clinical",
      "from": "BIOLOGIST",
      "to": "CLINICAL_LEAD",
      "labels": ["bio_report"]
    },
    {
      "id": "ch_tox_to_clinical",
      "from": "TOXICOLOGIST",
      "to": "CLINICAL_LEAD",
      "labels": ["tox_report"]
    },
    {
      "id": "ch_clinical_to_reg",
      "from": "CLINICAL_LEAD",
      "to": "REGULATORY_SPECIALIST",
      "labels": ["clinical_review"]
    },
    {
      "id": "ch_reg_to_director",
      "from": "REGULATORY_SPECIALIST",
      "to": "PROJECT_DIRECTOR",
      "labels": ["approval", "conditional_approval"]
    },
    {
      "id": "ch_director_to_chem",
      "from": "PROJECT_DIRECTOR",
      "to": "CHEMIST",
      "labels": ["final_decision", "retry", "abort"]
    },
    {
      "id": "ch_director_to_form",
      "from": "PROJECT_DIRECTOR",
      "to": "FORMULATION_SCIENTIST",
      "labels": ["retry", "done"]
    },
    {
      "id": "ch_reg_to_clinical",
      "from": "REGULATORY_SPECIALIST",
      "to": "CLINICAL_LEAD",
      "labels": ["hold_clearance", "no_hold"]
    },
    {
      "id": "ch_director_to_clinical",
      "from": "PROJECT_DIRECTOR",
      "to": "CLINICAL_LEAD",
      "labels": ["retry", "done"]
    },
    {
      "id": "ch_director_to_reg",
      "from": "PROJECT_DIRECTOR",
      "to": "REGULATORY_SPECIALIST",
      "labels": ["retry", "done"]
    }
  ]
}
\end{lstlisting}

\section{Pipeline Cost Across the Benchmark}
\label{app:pipeline-cost}

Table~\ref{tab:pipeline-cost-summary} reports aggregate pipeline cost across the 48-task benchmark.
We separate the one-time protocol-synthesis cost (Claude Opus 4.6 agentic loop, including IR generation, PlusCal writing, TLC verification, and counterexample-driven repair) from the per-execution runtime cost (gpt-5-mini, seven agents on average under Topology-monitored).

\begin{table}[h]
\centering
\caption{Aggregate pipeline cost across the 48-task benchmark. Synthesis = one-time agentic loop on Claude Opus 4.6 (IR generation, PlusCal writing, TLC verification, counterexample-driven repair). Runtime = mean per execution under Topology-monitored on gpt-5-mini.}
\label{tab:pipeline-cost-summary}
\small
\begin{tabular}{@{}lr@{}}
\toprule
Metric & Value \\
\midrule
\multicolumn{2}{@{}l}{\emph{Protocol synthesis (one-time, Claude Opus 4.6)}} \\
~~Tasks verified                       & 48/48 \\
~~Total repair attempts                & 29 (across 18 tasks) \\
~~Mean synthesis wall-clock per task   & $\sim$594\,s \\
~~Mean synthesis cost per task         & $\sim$\$4.33 \\
\midrule
\multicolumn{2}{@{}l}{\emph{Runtime per execution (gpt-5-mini, Topology-monitored)}} \\
~~Mean duration per execution          & 93\,s \\
~~Mean cost per execution              & $\sim$\$0.19 \\
\midrule
\multicolumn{2}{@{}l}{\emph{Sequential single-agent baseline (gpt-5-mini)}} \\
~~Mean duration per execution          & $\sim$304\,s \\
~~Mean cost per execution              & $\sim$\$0.10 \\
\bottomrule
\end{tabular}
\end{table}

Two patterns are worth noting.
First, TLC itself is not the bottleneck: model-checking completes in under 60\,s for every task (Section~4.2), so the agentic reasoning loop dominates the synthesis budget.
Second, the sequential baseline is cheaper per execution in dollar terms (a single agent avoids duplicated context and coordination tool calls), but is roughly $3\times$ slower in wall-clock and forfeits the formal coordination guarantees the verified protocol provides.
TraceFix amortizes its up-front protocol-synthesis cost across repeated runs of the verified protocol.

\section{Scoping; When Lock-and-Channel Coordination is the Right Model}
\label{app:scoping}

TraceFix targets the subset of multi-agent settings where coordination cannot be designed away: tasks that require shared mutable state with fine-grained synchronization.
The benchmark deliberately includes domains where this constraint is intrinsic.
Scenario 13 (Pharmaceutical Lab) models shared instruments such as HPLC and mass spectrometers; Scenario 15 (Semiconductor Fabrication) models shared processing chambers and metrology stations; Scenario 16 (CI/CD Pipeline) models shared build servers and deployment locks.
In each, two or more agents must access the same physical or virtual resource under mutual exclusion, and that constraint cannot be eliminated by architectural simplification.

Several other coordination paradigms work well outside this regime, and TraceFix does not compete with them.
\emph{Directed-acyclic-graph (DAG) orchestration} is appropriate when the dependency structure between subtasks is statically known and there is no contention; LangGraph and similar frameworks excel here.
\emph{Isolation-based} architectures, exemplified by Claude Code's per-agent git-worktree model, eliminate contention by giving each agent an independent copy of the shared state; this works elegantly for file-based code editing but does not generalize to physical resources (an HPLC instrument cannot be cloned per agent).
\emph{Sequential pipelines} and \emph{actor-model} architectures handle producer/consumer flows efficiently when each item is processed exactly once.

The lock-and-channel model TraceFix verifies is the appropriate abstraction precisely when these alternatives do not apply: multiple concurrent agents, shared mutable resources, fine-grained mutual exclusion, and bidirectional handshakes that cannot be flattened into a one-shot DAG.

\section{Per-Scenario Descriptions}
\label{app:scenario-descriptions}

The benchmark spans 16 scenarios with three difficulty tiers each (Easy / Medium / Hard, 48 tasks total).
Every scenario is paired with a simulation harness that exposes domain-specific tools, records per-checkpoint progress, and supports configurable fault injection (\texttt{--difficulty 0--3} for failure-injection rates; \texttt{--scenario N} for deterministic failures on the first $N$ decision-tool calls).
The harnesses compute the Avg Sim and Sim 100\% metrics reported in Section~4 and exercise the runtime's recovery paths under retryable errors (e.g., compile failures, instrument calibration failures, build-server timeouts).

\begin{itemize}[leftmargin=*]
\item \textbf{Scenario 1; Shared Codebase Development.} Developers modify shared code modules (auth, database, API) under exclusive access. Models concurrent feature development with module-level lock contention.
\item \textbf{Scenario 2; Research Paper Writing.} Researchers co-author paper sections (intro, methods, results); each section may be edited by only one researcher at a time. Models multi-author writing with section-level mutual exclusion.
\item \textbf{Scenario 3; Multi-Author Research Reports.} Researchers, a data analyst, a reviewer, and an editor-in-chief produce a paper with figures, peer review, and editorial assembly. Models hub-and-spoke writing pipelines with revision loops.
\item \textbf{Scenario 4; Frontend-Backend API Coordination.} A frontend developer, a backend developer, and a tester agree on an API contract before validating the integrated feature. Models paired-dev work with handshake and validation steps.
\item \textbf{Scenario 5; Medical Consultation.} An ER doctor coordinates with specialists (cardiology, pulmonology) and imaging to triage a patient. Models clinical decision-making with consultation requests.
\item \textbf{Scenario 6; Shared Utility File Conflict.} Developers each implement a feature touching a shared utility file; a reviewer must approve before merge. Models concurrent edits with review gates.
\item \textbf{Scenario 7; Document Co-authoring.} Writers each own a chapter but may edit each other's for consistency; an editor reviews the full document and may request revisions. Models cross-chapter co-authoring with iterative editing.
\item \textbf{Scenario 8; Backend-Database Schema Sync.} A backend developer, a database administrator, and a tester coordinate a schema migration that strictly precedes dependent backend deployment. Models ordered multi-stage deployment with cross-team dependencies.
\item \textbf{Scenario 9; Dining Philosophers.} Philosophers sit around a table sharing forks; each must hold both adjacent forks to eat, then put them down. Models the classical lock-order deadlock pattern.
\item \textbf{Scenario 10; Parallel Build.} Builders compile modules in parallel against a shared core library and shared type definitions; an integrator links the artifacts. Models build-system contention with sequential integration.
\item \textbf{Scenario 11; Flexible Manufacturing.} Technicians complete jobs on shared workstations (lathe, mill); when one is occupied, the technician switches to the other; a supervisor collects completion reports. Models adaptive scheduling under workstation contention.
\item \textbf{Scenario 12; Collaborative Kitchen.} Cooks share equipment (oven, stovetop, mixer) under exclusive access, with a head chef coordinating recipes. Models small-team service kitchens with fine-grained equipment scheduling.
\item \textbf{Scenario 13; Pharmaceutical Lab.} Lab technicians and a QC analyst share instruments (HPLC, mass spectrometer, balance), with a regulatory officer enforcing batch-record completeness. Models GMP-style laboratory workflows with regulatory checkpoints.
\item \textbf{Scenario 14; Drug Discovery Pipeline.} Researchers share cell cultures and assay equipment (plate reader, sequencer); a pipeline manager coordinates assay scheduling. Models high-throughput screening workflows with shared biological resources.
\item \textbf{Scenario 15; Semiconductor Fabrication.} Process engineers share fabrication chambers and metrology stations; a fab manager schedules tool access. Models cleanroom production with shared specialized equipment and strict ordering constraints.
\item \textbf{Scenario 16; CI/CD Pipeline.} Developers, CI runners, and a release manager share build servers and deployment locks. Models software release engineering with concurrent commits, build queues, and staged rollouts.
\end{itemize}

\end{document}